%% file: main.tex
\documentclass{article}

     \usepackage[final,nonatbib]{neurips_2022}

\usepackage[utf8]{inputenc} 
\usepackage[T1]{fontenc}    
\usepackage{hyperref}       
\usepackage{url}            
\usepackage{booktabs}       
\usepackage{amsfonts}       
\usepackage{nicefrac}       
\usepackage{microtype}      
\usepackage{xcolor}         
\usepackage{amsmath}
\usepackage{mathtools}
\usepackage{multirow}
\usepackage{bbm}
\usepackage{breqn}
\usepackage{dsfont}
\include{math_commands}

\usepackage{caption}
\usepackage{subcaption}

\DeclareMathOperator\erf{erf}

\title{FP8 Quantization: The Power of the Exponent}

\author{%
  Andrey Kuzmin\thanks{Equal contribution} , Mart Van Baalen\footnotemark[1] , Yuwei Ren, \\ \textbf{Markus Nagel, Jorn Peters, Tijmen Blankevoort} \\ 
  Qualcomm AI Research\thanks{Qualcomm AI Research is an initiative of Qualcomm Technologies, Inc.} \\
  \texttt{ \{akuzmin,mart,ren,markusn,jpeters,tijmen\}@qti.qualcomm.com }
}

\begin{document}

\maketitle

\begin{abstract}

When quantizing neural networks for efficient inference, low-bit integers  are the go-to format for efficiency. However, low-bit floating point numbers have an extra degree of freedom, assigning some bits to work on an exponential scale instead. This paper in-depth investigates this benefit of the floating point format for neural network inference. We detail the choices that can be made for the FP8 format, including the important choice of the number of bits for the mantissa and exponent, and show analytically in which settings these choices give better performance. Then we show how these findings translate to real networks, provide an efficient implementation for FP8 simulation, and a new algorithm that enables the learning of both the scale parameters and number of exponent bits in the FP8 format. Our chief conclusion is that when doing post-training quantization for a wide range of networks, the FP8 format is better than INT8 in terms of accuracy, and the choice of the number of exponent bits is driven by the severity of outliers in the network. We also conduct experiments with quantization-aware training where the difference in formats disappears as the network is trained to reduce the effect of outliers\footnote[1]{Code will be made available at \url{https://github.com/Qualcomm-AI-research/FP8-quantization}}. 

\end{abstract}


\input{introduction}

\input{Background}
\input{quantization_error_analysis}

\input{quantization_method}

\input{experiments}
\input{related_work}

\input{limitations}
\input{conclusion}

\section*{Acknowledgments}
We would like to thank Christos Louizos and Yin Huang for helpful discussions and valuable feedback.

\bibliography{references}
\bibliographystyle{plain}

\newpage
\input{appendix}

\end{document}

%% file: math_commands.tex
\usepackage{amsmath,amsfonts,bm}

\newcommand{\quant}[1]{{#1^{(q)}}}
\newcommand{\intt}[1]{{#1^{(int)}}}

\def\bhat{{\widehat{b}}}

\def\eqref#1{equation~\ref{#1}}

\def\1{\bm{1}}

\def\vx{{\bm{x}}}

\def\mW{{\bm{W}}}
\def\mX{{\bm{X}}}
\def\mY{{\bm{Y}}}

\DeclareMathAlphabet{\mathsfit}{\encodingdefault}{\sfdefault}{m}{sl}
\SetMathAlphabet{\mathsfit}{bold}{\encodingdefault}{\sfdefault}{bx}{n}

\DeclareMathOperator*{\argmin}{arg\,min}

\usepackage{amsthm}

\theoremstyle{definition}

\theoremstyle{remark}

%% file: introduction.tex
\section{Introduction}

Neural network quantization is one of the most effective ways to improve the efficiency of neural networks.
Quantization allows weights and activations to be represented in low bit-width formats, e.g. 8 bit integers (INT8). When executing networks on any device, this leads to a reduction in data movement and enables the use of low bit-width computations, resulting in significantly faster inference and lower energy consumption.

Roughly speaking, values in neural networks are represented in either integer (INT) or floating-point formats (FP). Most quantization research has gone into low-bit integer formats such as INT8 and INT4, as the corresponding hardware for this is widely available. However, some research suggests there might be benefits to using low bit-width floating-point formats for neural network computations~\cite{sun2019_HFP8}.

In this paper, we set out to thoroughly investigate the potential benefits of the floating point format for neural network inference. We show that floating point numbers add an extra dimension on top of the INT8 format that makes outliers in the quantized distribution less harmful to the overall performance. 
We investigate the effect of the quantization formats on neural network quantization on three levels: 
1) Analytically for several common data and weight distributions, 2) practically in INT8 and FP8 post-training quantization (PTQ) settings, and 3) in quantization-aware training (QAT) settings with both INT8 and different FP8 formats.
We will show there is a strong agreement between our theoretical results and our practical results on real networks.

In order to compare the two formats, we introduce a novel floating-point quantization simulation implementation that enables us to quickly run PTQ and QAT experiments with many floating point formats.
Furthermore, this quantizer enables us to learn the FP8 exponent bias value, as well as the best trade-off between the number of bits used for the exponent and the mantissa, making good use of the flexibility the floating-point format provides and providing a way to learn this automatically without manual intervention.  
Our conclusion from the study presented in this paper is that floating-point quantization can be better than integer quantization for inference, but it needs a carefully tuned bias term and correct bit-settings between the exponent and mantissa.

%% file: background.tex
\section{Background}

\subsection{Integer quantization}\label{sec:intquant}
Quantization is often applied to enable efficient integer matrix multiplications instead of expensive 32-bit floating point (FP32) multiplications. 
A matrix $\mX\in \mathbb{R}^{m\times n}$ is quantized to an integer matrix $\intt{\mX}$ with an associated scale $s$ as follows:     $\intt{\mX}=\text{clip}\left(\left\lfloor \frac{\mX}{s}\right\rceil, x_{min}, x_{max}\right)$,
where $\lfloor \cdot \rceil$ indicates the round-to-nearest operation, and $\text{clip}(\cdot, \cdot, \cdot)$ indicates the element-wise clipping operation, which ensures that $\intt{\mathbf{X}}$ can be represented in the chosen bit-width \cite{whitepaper}.
The \emph{dequantization} operation then produces a matrix $\quant{\mathbf{X}}$ which is a quantized approximation of the input matrix $\mX$: $\quant{\mX}=s\intt{\mX}\approx\mX.$
This allows the use of efficient integer matrix multiplication~\cite{whitepaper}. For a matrix $\mY\in \mathbb{R}^{n\times k}$,
\begin{equation}
\mX\mY\approx\quant{\mX}\quant{\mY}=s_x s_y \intt{\mX}\intt{\mY}.    
\end{equation}

\begin{figure}[t]
    \centering
    \includegraphics[width=\textwidth]{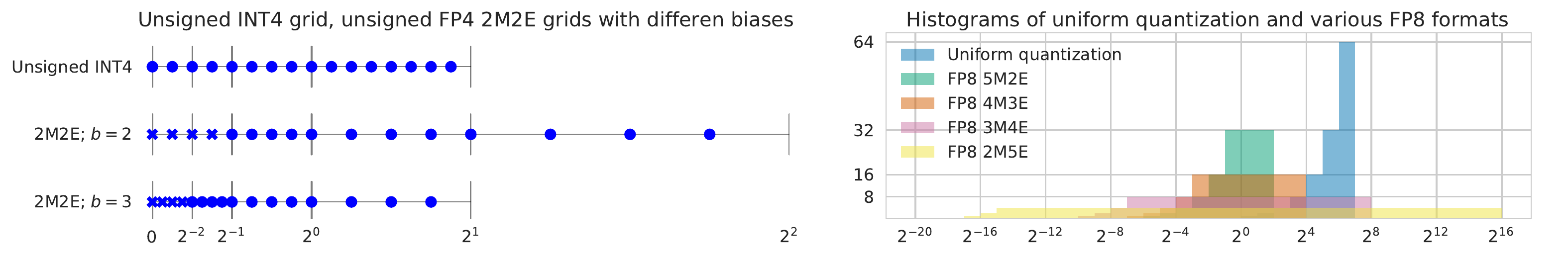}
    \caption{
        \textbf{Left:} Unsigned INT4 quantization compared to unsigned FP4 2M2E quantization with two different biases. 
        Subnormal ranges are marked with crosses, normal range with circles.
        Compared to uniform quantization, FP quantization has larger dynamic range, but less precision for larger values.
        \textbf{Right}: Histograms of the positive values representable by 8 bit signed integers and signed 8 bit floating point formats.
        Histogram bins are log-spaced, to show the relation between dynamic range (i.e. number of bins) and precision (i.e. values per bin) between different formats.
        FP8 formats allow values with larger dynamic range, at the expense of precision.
        Intuitively, changing the (floating point) bias value for each distribution would move these histograms to the left or right without changing their shape.
    }
    \label{fig:fp8_grid}
\end{figure}
\subsection{Floating point number system}
A floating point number set $F \subset \mathbb{R}$ is a set whose elements are defined as follows:
\begin{equation}
f = (-1)^s 2^{p-b} \left( 1 + \frac{d_1}{2} + \frac{d_2}{2^2} + \cdots \frac{d_m}{2^m} \right),
\label{eq:fp_system}
\end{equation}
where $s\in\{0,1\}$ is the sign bit, $d_i \in \{0,1\}$ is the $m$-bit mantissa, $p\in\mathbb{Z};0\leq p < 2^e$ is the $e$-bit exponent, and $b$ is an integer \emph{exponent bias}, commonly defined to be $2^{e-1}$.



Floating point numbers can be seen as a uniform $m$-bit grid between two consecutive (integer) powers of two $2^a, 2^{a+1}$.
The distance between grid points in the range $[2^a, 2^{a+1}]$ is $2^{a-m}$.
Increasing the number of mantissa bits thus increases the number of grid points in each range $[2^a, 2^{a+1}]$.
In the definition provided earlier in this section, the number of ranges $[2^a, 2^{a+1}]$ that can be represented by a floating point number system, is determined by the number of exponent bits $e$.
Increasing the number of exponent bits thus increases the dynamic range (i.e. ratio between largest and smallest non-zero value) of values that can be represented.
Any fixed bit-width floating point number system must make a trade-off between the dynamic range of representable values ($e$), and the precision ($m$).
For example, IEEE-754 32-bit `full-precision' floating point numbers (FP32) use 1 sign bit, 23 mantissa bits, and 8 exponent bits.
The resulting effect is that, compared to integer formats, floating point formats have more precision close to zero, as the ranges $[2^a, 2^{a+1}]$ will be smaller for lower values of $a$, and less precision away from 0. This is visualized in the left plot in  Fig.~\ref{fig:fp8_grid}. 
Intuitively, they are a better match for peaked distributions like Gaussians that have more density around 0, and a better fit for distributions with large tails and outliers like the Student's t-distribution. 
The right plot in Fig.~\ref{fig:fp8_grid} shows the different distributions of values for various FP8 formats.

Note that this definition does not allow for a representation of 0.
To allow 0 to be represented, the exponent value $p=0$ is reserved to indicate \emph{subnormal} numbers.
In this case, the exponent value is implicitly set to $1$, and $f = (-1)^s 2^{1-b} \left(0 + \frac{d_1}{2} + \frac{d_2}{2^2} + \cdots \frac{d_m}{2^m} \right).$
Besides allowing the exact representation of 0, subnormal numbers also allow a graceful representation of values close to 0.
See Fig.~\ref{fig:fp8_grid} for an intuition behind this.
In Section~\ref{sec:fp8_quantsim} we show (de)quantization operations analogous to those of INT8.

\subsection{Assumptions and extensions}
In this work we make a number of assumptions and consider several extensions of the standard floating point format.
We assume that the exponent bias $b$ is not restricted to the value $2^{e-1}$. Instead, we introduce a (per-tensor or per-channel) quantization scale $\gamma$, similar to the quantization scale used in integer quantization \cite{whitepaper, krishnamoorthi}.
Since there is no standard allocation of mantissa and exponent bits for 8 bit floating point formats, we consider various allocations of mantissa and exponent bits, as we assume that the choice of trade-off between precision and dynamic range will have more impact for lower floating point bit-widths. 
We use the notation xMyE to denote a format with x mantissa bits and y exponent bits. 
Specifically, we consider the formats 5M2E, 4M3E, 3M4E and 2M5E.


%% file: quantization_error_analysis.tex
\section{Expected quantization error}\label{sec:expected_quantization_error}
\begin{figure}[t]
\centering
\begin{tabular}{c}
\includegraphics[width=13.0cm]{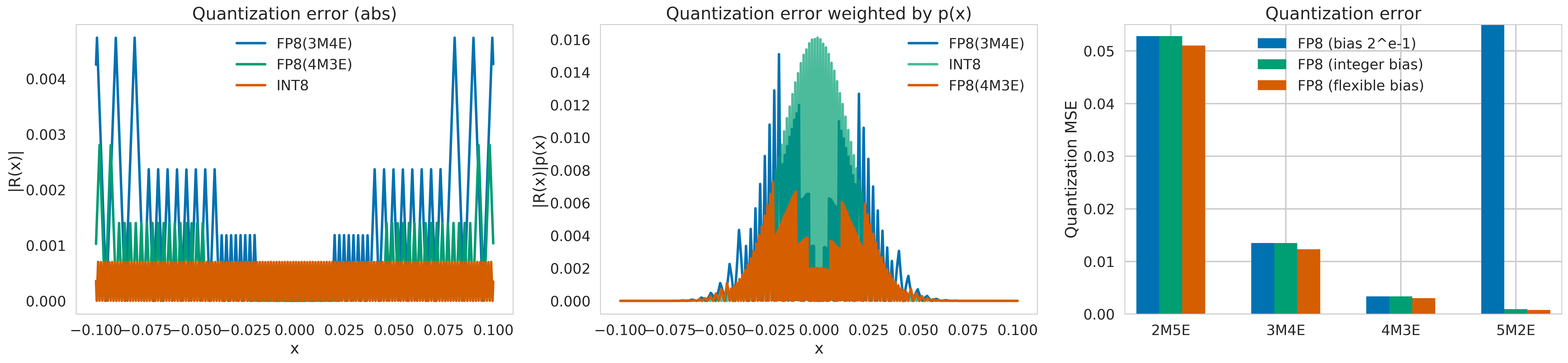}
\end{tabular}
\caption{(left) Rounding error for quantization grids: FP8 with 4-bit exponent, FP8 with 3-bit exponent, INT8. (middle) The rounding error weighted by the probability density function from the first column. (right) Exponent bias ablation for a zero mean Gaussian distribution for FP8 format. We consider three versions of the format: fixed format with the bias $2^{e-1}$, integer bias, FP32 scale.}
\label{fig:distr_grid_mse}
\end{figure}
In this section we perform an analytical analysis of the FP8 and INT8 formats and show that formats with more exponent bits perform better when outliers are a factor. This theoretical result lays a foundation for the comparison between the formats that is predictive of when to use the FP8 format and how many exponent bits to use. 

We investigate the error induced by quantizing values drawn from several insightful distributions, considering the expected MSE of these values as it has been shown to be indicative of the final loss in a neural network \cite{adaround}.


\subsection{Expected quantization error}

Given a quantization grid $\alpha=\{\alpha_1, \alpha_2, \dots \alpha_k\}$, we can define the quantization operation $Q_{\alpha}(w)$ and the corresponding quantization error $R_{\alpha}(w)$:
\begin{align}\label{eq:fp8_nearest}
Q_{\alpha}(w) = \alpha_i, \text{ } i=\argmin_{i}|w-\alpha_i|, & & R_{\alpha}(w) = Q_{\alpha}(w) - w.
\end{align}



Examples of the rounding error function for the 3M4E, 4M3E, and INT8 formats are shown in fig.~\ref{fig:distr_grid_mse}(left).
We model neural network weights as a random variable $W \sim p_w(w)$. The expected value of the quantization MSE can be expressed as follows: 
\begin{equation}
\label{eq:rounding_error}
\begin{split}
{\mathbb E}(R_{\alpha}(W))^2=
\int\limits_{\alpha_{min}}^{\alpha_{max}}R_{\alpha}^2(w)p_wdw + 
\int\limits_{-\infty}^{\alpha_{min}}(w-\alpha_{min})^2p_wdw + \int\limits_{\alpha_{max}}^{\infty}(\alpha_{max}-w)^2p_wdw,
\end{split}
\end{equation}
where $\alpha_{min}=\min_{i}{\alpha_i}$ and $\alpha_{max}=\max_{i}{\alpha_i}$ are the limits of the quantization range. The first term corresponds to the rounding error, while the remaining two terms correspond to the clipping error. We will use this formulation to analytically compute the quantization errors for different distributions. This is done by splitting the integration into sub-intervals corresponding to each point of the quantization grid. We present further details of this procedure the appendix~\ref{sec:integral_analyt}. An example of the rounding error function weighted by the probability density is given in fig.~\ref{fig:distr_grid_mse}(middle).

\begin{figure}[t]
\centering
\begin{small}
\begin{tabular}{c}
\includegraphics[width=13.0cm]{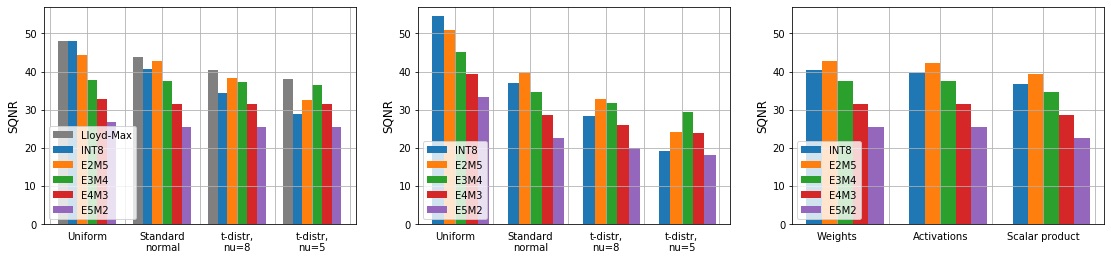} 
\end{tabular} 
\end{small}
\caption{(left) Quantization expected SQNR for different distributions and different formats including Lloyd-Max non-uniform quantizer~\cite{lloyd1982least,max1960quantizing}. The Student's-t distribution is clipped at $[-100,100]$ to avoid an unbounded error. (middle) scalar product expected SQNR, (right) Expected SQNR for a Resnet18 layer.}
\label{fig:rounding_error}
\end{figure}

\subsection{Scalar product quantization error}
\label{sec:scalar_product}
We can now model the expected output error for a quantized scalar product, the essential building-block for both fully-connected and convolutional layers. In order to model the activation values, we introduce a random variable $X \sim p_x(x)$ and the corresponding quantization grid $\beta=\{\beta_1, \beta_2, \dots \beta_k\}$, $\beta_i \in \mathbb{R}$. 
While computing the scalar product of $X$ and $W$, we only consider the error incurred by quantization of the inputs, as the error incurred by quantization of the output $Y$is identical to the input quantization error as described in the previous section.

We denote the output tensor $Y=WX$, and its quantized version as $Y_Q$. The quantization error $\Delta Y$ then can be expressed as
\begin{equation}
\begin{split}
\Delta Y = Y_Q - Y = Q_{\alpha}(W)Q_{\beta}(X)-WX=WR_{\beta}(X)+XR_{\alpha}(W)+R_{\beta}(X)R_{\alpha}(W). 
\end{split}
\end{equation}



The sample mean value $\overline{\Delta Y^2}$ corresponds to the mean error of the scalar product between two vectors $\mathbf{W} \in \mathbb{R}^n$ and $\mathbf{X} \in \mathbb{R}^n$, where $W_i \sim p_w(w)$, $X_i \sim p_x(x)$ are weights and activations samples: 
\begin{equation}
\overline{\Delta Y^2} = \frac{1}{n} \left[ \langle Q(\mathbf{W}),Q(\mathbf{X}) \rangle - \langle \mathbf{W},\mathbf{X} \rangle \right]^2.
\end{equation}

Computing ${\mathbb E} (\Delta Y^2)$ for different distributions and quantization grids allows us to analyze how they would work in practical scenarios. In many cases, the expected output MSE can be approximated by the following weighted sum of the rounded errors on $W$ and $X$:
\begin{equation}
\label{eq:scalar_error_approx}
{\mathbb E} (\Delta Y^2) \approx {\mathbb E}(R_{\alpha}(W))^2\int\limits_{-\infty}^{\infty} x^2 p_x(x) dx   + {\mathbb E}(R_{\beta}(X))^2\int\limits_{-\infty}^{\infty} w^2 p_w(w) dw.     
\end{equation}

We see that the output error mainly depends on the quantization error of the inputs and the spread of the distributions of the inputs. We describe the full procedure for computing ${\mathbb E} (\Delta Y^2)$ and further analysis in appendix~\ref{sec:appendix_scalar_product}. 


\paragraph{Discussion}

We use these formulations to analytically compute the quantization errors for several distributions. We consider the uniform distribution, Gaussian distribution and the student-t distribution. The latter is essentially a more heavy-tailed Gaussian, allowing us to model what happens when outliers are added into the picture. 

It is important to choose the bias term in the FP8 format properly. We see in fig. \ref{fig:distr_grid_mse}(right) that the standard fixed format of setting the bias to $2^{e-1}$ can fail. This happens when values are either clipped too much, or not enough grid-points are used as the representable range is too big. We also see that having an integer bias performs worse than having a floating-point bias. This is similar to what happens in INT8 quantization, where setting the scale parameter correctly can have a large impact on the network's performance \cite{whitepaper}. As the bias shifts the dynamic range, this effect is less strong for the FP8 formats with a high amount of exponent bits that have a high dynamic range. For lower exponent-bits, one would preferably use a per-channel bias/scale parameter, similar to what is done in per-channel quantization \cite{krishnamoorthi}. Further justification for this can be found in section~\ref{sec:ptq}.

 We also analyze the effect of different settings of the number of mantissa and exponent bits. The results of this are presented in fig.~\ref{fig:rounding_error}. We observe a clear pattern. For a uniform distribution, the INT8 format performs the best. For the Gaussian distribution, the format with 2 exponent bits performs best. Then, when increasing the relative outliers by decreasing the degrees of freedom in the Student's t-distribution, the best format tends towards having more exponent bits. 

In neural networks, most weights and activations can be modeled relatively well by a Gaussian distribution. So based on this analysis, we would expect the 5M2E format to work the best for most well-behaved layers with little outliers. Specifically in the weights outliers are infrequent, as the weights are often explicitly, and otherwise implicitly regularized~\cite{zhang2021understanding}. 
For example, we consider a layer of Resnet18 model pre-trained on ImageNet. 
We fit Gaussian distributions in the weights and activations sample, and compute the expected MSE analytically in fig.~\ref{fig:rounding_error} (right). We see that the 5M2E format works the best in this case. A more detailed per-layer study is given in appendix~\ref{sec:appendix_conv_layer_ablation}.

However, the stronger the outliers are in the distribution, the more the FP8 format with a higher number of bits will help with its quantized representation. We would thus predict that for networks with severe outliers, formats like M4E3 would perform better. 
We will show in section \ref{sec:experiments} that these predictions hold for real networks, and that networks that are known to have larger activation outliers such as transformers benefit the most from more exponent bits.

%% file: quantization_method.tex
\section{FP8 quantization simulation}\label{sec:fp8_quantsim}

In the previous section our analysis showed that FP8 quantization can theoretically yield better tensor reconstruction MSE than INT8 quantization, but careful selection of the division between exponent and mantissa bits, as well as the value of the exponent bias are crucial.
We investigate whether these findings can be extended to FP8 quantization of neural networks. 
In order to do so, we need an efficient simulation of FP8 quantization.
Methods commonly used for FP8 quantization are either too slow, e.g. nearest grid point search or bit-masking \cite{Huang2021_8bit_flex_fp}, or too complicated, e.g. custom CUDA kernels \cite{sun2019_HFP8}, for quick experimentation.

In this section we introduce a novel method of FP8 quantization simulation.
This method can easily be implemented in common deep learning frameworks.
An added benefit of this method is that it exposes the parameters of the FP8 quantizer (i.e., the number of mantissa/exponent bits and the exponent bias), thus allowing the parameters to be learned by back-propagation.

\subsection{FP8 simulated quantization formulation}
In our FP8 quantizer we exploit the fact that FP8 quantization can be seen as the union of $m$-bit uniform quantization grids between consecutive integer powers of two $[2^{p-b}, 2^{p+1-b}]$.
This means we can simulate FP8 quantization of an input vector\footnote[2]{This method extends readily to matrices or higher-dimensional tensors.} $\vx$ using the same method as for simulation of uniform quantization as described in Section~\ref{sec:intquant}, with the distinction that each element $x_i$ in $\vx$ has its own associated scale $s_i$:
\begin{equation}\label{eq:rp8_round_to_nearest}
    \quant{x_i}=s_i \left\lfloor \frac{x_i}{s_i} \right\rceil,
\end{equation}
where $\quant{x_i}$ denotes $x_i$ quantized to floating point.
The scale $s_i$ depends on the number of mantissa bits $m$ and the range $[2^{p-b}, 2^{p+1-b})$ in which $x_i$ falls.
This is given by $\log_2 s_i=p_i=  \left\lfloor \log_2 |x_i| \right\rfloor - m$.

To ensure that $\quant{x_i}$ can be represented given $m$, $e$ and $b$, both $\quant{x_i}$ and $s_i$ need to be clipped.
Values of $\quant{x_i}$ greater than maximum value $c$ or smaller than $-c$ are clipped $c$ and $-c$ respectively, where $c = (2 - 2^{-m})2^{2^e-b-1}$ is the largest representable value for a given floating point format.
Since $2^{1-b-m}$ is the smallest representable value, values of $p_i$ smaller than $1-b-m$ are clipped to $1-b-m$.



\begin{figure}[h]
    \centering
        \begin{small}
            \includegraphics[width=0.75\textwidth]{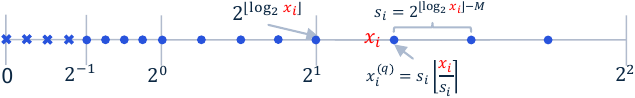}
        \end{small}
    \caption{Graphic depiction of the FP8 quantizer.}
    \label{fig:fp8quantizer}
\end{figure}

Note that this approach is identical to the quantization operation defined in Eq.~\ref{eq:fp8_nearest}, provided that the rounding mode in Eq.~\ref{eq:rp8_round_to_nearest} matches the tie-breaking procedure in Eq.~\ref{eq:fp8_nearest} for numbers equidistant to two numbers in $F$.
See Fig.~\ref{fig:fp8quantizer} for an intuition.

In case the scaling factor $\gamma\neq 1$ this needs to be reflected in $p_i$.
In order to accommodate the scaling factor, we first fold it into a reparameterized bias value $\bhat=b-\log_2\gamma$. 
We then compute $p_i$ as follows:
\begin{equation}
    p_i = 
    \begin{dcases}
        \left\lfloor \log_2|x_i| + \bhat \right\rfloor - \bhat - \lfloor m\rceil, & \text{if } \left\lfloor \log_2 |x_i| +\bhat \right\rfloor > 1 \\
        1-\bhat-\lfloor m\rceil, & \text{otherwise}.
    \end{dcases}
    \label{eq:pi_with_scale}
\end{equation}

\subsection{Quantization-aware training with FP8}
To enable QAT using this quantizer, we need to make a few changes.
First, to allow gradients to flow through each step of the quantizer, we use the straight-through estimator (STE) for gradients of non-differentiable rounding operations. 
Second, we find that learning the maximum clipping value $c$ instead of $\bhat$ improves training stability.
$\bhat$ can be found from $c$ as follows: $\bhat=2^e-\log_2 c - \log_2 (2-2^{-m}) - 1$.
Lastly, we treat $\left\lfloor \log_2|x_i| + \bhat \right\rfloor$ as a constant that receives no gradient. 
This prevents the (sometimes extremely large) gradients of this operation w.r.t. $x_i$ to propagate backwards.
The result of this is that $\vx$ receives the `straight-through' gradient for the full quantization procedure, i.e.  $\frac{\partial}{\partial x_i}F(x_i, m, c)=1$, where $F(\cdot,\cdot,\cdot)$ denotes the FP8 quantizer.

\subsection{Toy experiment: Learning minimal MSE on common distributions}
\begin{figure}[h]
    \centering
    \includegraphics[width=\textwidth]{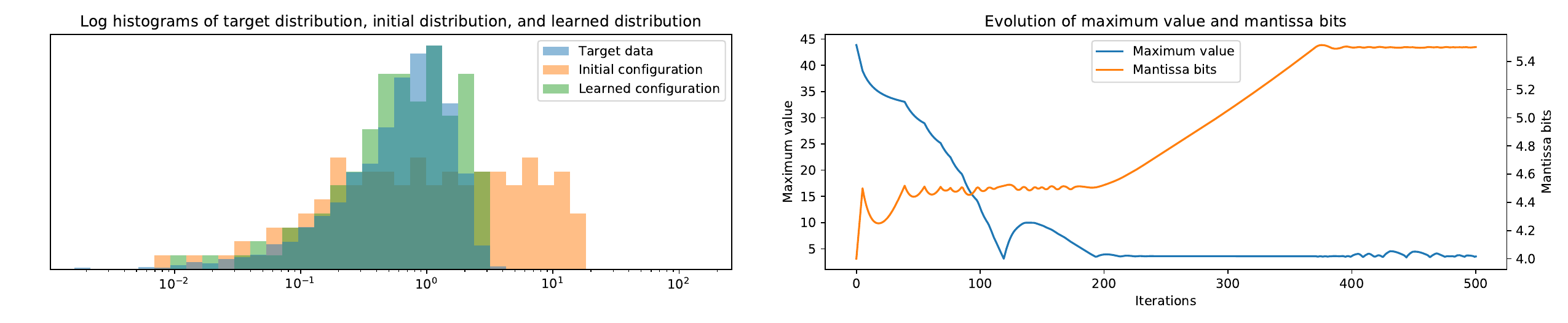}
    \caption{\textbf{Left}: Histograms of target data, and representable values of the initial format (3M4E) and the learned format (5M2E). \textbf{Right}: evolution of maximum value and number of mantissa bits. After roughly 350 iterations the number of mantissa bits starts oscillating around a value of 5.5.}
    \label{fig:fp8_learning_toyexp}
\end{figure}

To investigate whether our quantizer can indeed learn the maximum value $c$ and number of mantissa bits $m$, we run a toy experiment.
We sample $10^5$ values from $\mathcal{N}(0, 1)$, and initialize an FP8 quantizer with 3M4E and a bias of 8, which corresponds to maximum value of $c=240$.
We then use SGD to learn the values of $c$ and $m$ that minimize MSE on the reconstruction loss: $\mathcal{L}(M,c)=\frac{1}{N}\sum_i \left( x_i - F(x_i, m, c) \right)^2$.
After 500 iterations, $c$ has converged to 4.35 and $m$ oscillates around 5.5.
The oscillation behavior can be explained by the fact that MSE can be further minimized by increasing precision through higher $m$, however increasing $m$ to 6 yields a uniform quantizer, which results in higher MSE.
Similar behavior was observed in uniform quantization by \cite{nagel2022oscillations}.
A line search shows that indeed $m=5$ and $c=4.37$ minimize the MSE on the target data, meaning our algorithm found the optimal values in this example. 
See Fig.~\ref{fig:fp8_learning_toyexp} for details.

%% file: experiments.tex

\section{Experiments}
\label{sec:experiments}
In this section, we aim to empirically validate the analytical findings from section \ref{sec:expected_quantization_error} on full neural networks.
We use our analytical results to investigate various FP8 formats for weight and activation tensors in neural networks, and show that
Finally, we perform quantization-aware training (QAT) on FP8 quantized models.

\subsection{Baselines and experimental setup}\label{sec:baselines}
We run our experiments on a variety of different tasks, datasets, and models: We experiment on ResNet18~\cite{resnet}, MobileNetV2~\cite{mobilenetv2}, and ViT~\cite{vit} for ImageNet classification~\cite{imagenet}; BERT-base~\cite{bert} for language understanding on the GLUE benchmark~\cite{wang2019glue}; HRNet~\cite{Sun_2019_CVPR} for semantic segmentation on the Cityscapes dataset~\cite{Cordts2016Cityscapes}; DeepLabV3~\cite{deeplabv3} for semantic segmentation on the Pascal VOC dataset~\cite{everingham2015pascal}; and SalsaNext~\cite{salsanext} for LIDAR point cloud segmentation on the SemanticKITTI dataset~\cite{behley2019iccv}.
For each model, we report the 32-bit floating point (FP32) results on the same model instance (trained weights and code base) used in our other experiments.

As baselines we compare against models quantized using INT8 quantization.
Following \cite{nagel2022oscillations} we do not apply batch normalization folding, and re-estimate the batch normalization statistics (running mean and variance) before final validation, as this improved results for every model we considered.
For each model, we consider several methods for range estimation separately for weights and activations, per-channel and per-tensor range estimation for weights.
Results reported are those for the best combination of range estimation setting and per-channel or per-tensor range estimation.
Our code is written in PyTorch and all our experiments are performed using NVIDIA Tesla V100 and A100 GPUs.

\subsection{Post-training quantization results}\label{sec:ptq}

We compare post-training quantization using INT8 and FP8 quantization. 
To allow an apples-to-apples comparison, we do not use any methods to improve post-training quantization that have been developed in the past years (e.g. those mentioned in \cite{whitepaper}), other than the methods described in Section~\ref{sec:baselines}.

For our FP8 results, we report the best fully fixed format, i.e. the same combination of $m, e$ and $b$ used for the full network, the best flexible bias format, i.e. $m$ and $e$ fixed for the full network, but $\bhat$ applied per channel or tensor, whichever yields best results, and a fully flexible format where $m$ and $e$ are set for each tensor, and $\bhat$ is set for each channel, to minimize MSE.
The full procedure for this approach is detailed in Section~\ref{sec:mse_mantissa_bias}.
For an example setting, see Figures~\ref{fig:rn18sqnr} and ~\ref{fig:bertsqnr} for an example of FP8 formats that minimize MSE for each tensor in ResNet18 and BERT, respectively.

\paragraph{Discussion} PTQ results can be found in Table~\ref{tab:ptq}.
In this table we can see that networks with large activation outliers (ViT, Bert, SalsaNext and HRNet) show better results for formats with more exponent bits, and thus a larger dynamic range.
Furthermore, we see that, as predicted in Section~\ref{sec:integral_analyt}, formats with more mantissa bits are better able to represent weight and activation tensors in convolutional neural networks.
However, since increasing the number of mantissa bits implies reducing dynamic range, finding the right value for the bias for each channel is important.
The full set of results for our PTQ experiments is detailed in Table~\ref{tab:ptqfull} in Appendix Section~\ref{sec:appendixptq}.


Lastly, we see that fully flexible formats only sometimes outperform fixed $m/e$ formats, with slim improvements when they do.
This is surprising, as a more flexible format should be able to better fit a network's tensors.
We attribute this discrepancy to our local greedy method for assigning $m, e$ and $\bhat$, which may not find settings that are optimal globally.



\input{tables/ptq}

\subsection{Quantization-aware training}
We investigate whether we can improve on the FP8 PTQ results by performing training with FP8 quantization in mind (quantization-aware training, QAT).
We compare against INT8 QAT, which has previously been shown to result in INT8 models with accuracy near the original FP32 models.

We consider three different initializations based on our PTQ results from Table~\ref{tab:ptq}: best fixed format, best flexible bias format and best fully flexible format.
For each of these initializations we perform QAT with the format fixed at initialization, with learnable maximum value $c$ (cf. range learning in INT8 QAT), and with learning both $c$ and the number of mantissa bits $m$.
We run experiments with various learning rates for model and quantization parameters, as well as per-tensor and per-channel quantization, and report results for the best learning setup.
We train our models for 20 epochs and use Adam for the model parameters and SGD for the quantization parameters.
Results of these experiments can be found in Table~\ref{tab:qat}.
Full experimental details and results can be found in Appendix~\ref{sec:expdetail}.

\begin{figure}[h]
    \centering
    \includegraphics[width=0.8\textwidth]{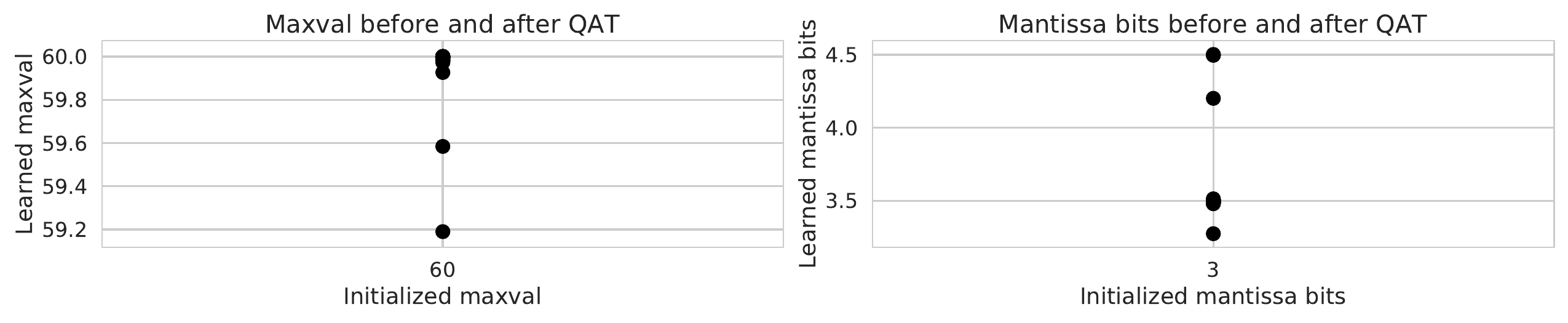}
    \caption{Values of maximum value $c$ and number of mantissa bits $m$ before and after training. In this experiment, maximum value was initialized to $c=60$, and number of mantissa bits was initialized to $m=3$. }\label{fig:fp8_before_after}
\end{figure}

\paragraph{Discussion} 
From these results, we see that QAT can improve performance of FP8 quantized models.
We also see that, for ResNet18 and MobileNetV2 learning both maximum value $c$ and number of exponent bits $m$ improves results compared to just learning weights, for fully fixed and flexible bias formats.
See also Figure~\ref{fig:fp8_before_after} for an example of how initialized $c$ and $m$ differ from learned $c$ and $m$ after 20 epochs of training.
However, this difference disappears for fully flexible PTQ initialization, and learning $c$ and $m$ slightly harms performance for BERT, although the difference is too small to be significant.

Generally, we see that QAT reduces the accuracy gap between different formats, as the weights of the network can adapt to the distributions the quantizers represent.


\input{tables/qat}

%% file: tables/ptq.tex
\begin{table}[]
    \centering
    \begin{tabular}{l|r r|r r r r r}
        Model & FP32 & INT8 & \multicolumn{2}{c}{Best fixed} &  \multicolumn{2}{c}{Best flex bias} &  Best flexible \\
        && & Format & Score & Format & Score & Score \\
        \hline
        ResNet18 &      69.72 &   69.55 &   3M4E9B &   68.52 & 5M2E & 69.66 & 69.64 \\
        MobileNetV2 &   71.70 &   70.94 &   4M3E3B  &   68.49 & 5M2E & 71.06 & 71.28 \\
        ViT &           77.75 &   76.39\  &   3M4E8B  &   77.37 & 4M3E & 77.71 & 77.62 \\
        BERT-base &     83.06   &   71.03   &   3M4E9B  &   83.08   & 3M4E & 82.80   & 83.02 \\
        SalsaNext &     55.80   &   54.22   &   4M3E3B  &   55.79   & 4M3E & 55.80   & 55.02 \\
        DeepLabV3 &     72.91   &   71.24   &   3M4E8B  &   9.42    & 5M2E & 72.58   & 72.72 \\   
        HRNet &         81.05   &   80.93   &   3M4E8B  &   80.71   & 4M3E & 81.04   & 81.06 \\
    \end{tabular} \vspace{.1cm}
    \caption{PTQ results with for fixed formats, flexible bias formats, and fully flexible formats, compared to original model (FP32) and INT8 results.}
    \label{tab:ptq}
\end{table}

%% file: tables/qat.tex
\begin{table}[]
    \centering
    \begin{tabular}{l|r r | r r r }
        Model               & FP32    & INT8    & Best fixed & Best flex bias &  Best flexible \\
                            &         &         &  accuracy  &  accuracy      & accuracy \\
        \hline
        ResNet18 PTQ        & \multirow{4}{*}{69.72} & 69.55 & 68.44 & 69.66 & 69.64 \\
        ResNet18 QAT        &                        & 70.43 & 69.16 & 69.82 & 70.28        \\
        + $c$ learning  &                        &       & 69.30 & 69.97 & 70.15        \\
        + $m$ learning      &                        &       & 70.10 & 70.25 & 70.24        \\
        \hline
        MobileNetV2 PTQ        & \multirow{4}{*}{71.70} & 70.94 & 68.49 & 71.06 & 71.28 \\
        MobileNetV2 QAT        &                        & 71.82 & 71.03 & 71.54 &  71.60             \\
        + $c$ learning      &                        &       & 70.93 & 71.67 &  71.66       \\
        + $m$ learning      &                        &       & 71.34 &  71.69    &  71.74 \\
        \hline
        BERT PTQ        & \multirow{4}{*}{83.06} & 71.03 & 83.08 & 82.80 & 83.02 \\
        BERT QAT        &                        & 83.26 & 83.56 & 83.70 & 83.17             \\
        + $c$ learning      &                        &       & 83.89 & 83.86 & 83.72       \\
        + $m$ learning      &                        &       & 83.85 & 83.84 & 83.70 \\
    \end{tabular} \vspace{.1cm}
    \caption{QAT results with for fixed formats, flexible bias formats, and fully flexible formats, compared to original model (FP32) and INT8 QAT results.}
    \label{tab:qat}
\end{table}

%% file: related_work.tex
\section{Related work}

\paragraph{Integer quantization} Integer quantization, sometimes also called fixed-point quantization, has been an active field of research since neural networks became more popular and the need for efficient inference became important. On a high level there are two lines of work, post-training quantization (PTQ) \cite{lin2016icml, dfq, dong2019hawq, bannerposttraining, huaweiquant, zeroshotquant, adaround, brecq}
and quantization-aware training (QAT) \cite{Gupta2015, jacob2018cvpr, dorefa, pact2018, louizos2018relaxed, lsq, tqt, lsq+, differentiablequantization, nagel2022oscillations}.
For more details we refer the interested read to the surveys of \cite{krishnamoorthi2018quantizing, whitepaper, gholami2021survey}.
Our approach to learning the FP8 configuration is inspired by \cite{lsq} which assume the straight through estimator (STE) \cite{bengio_estimating_2013} on the rounding operation in order to derive a gradient for the scaling factor of the fixed point format. \cite{differentiablequantization} extends this to jointly learn the bit width and scaling factor, which we explore for learning the trade-off between the number mantissa and exponent bits.


\paragraph{Floating point formats}
The accuracy and stability of floating point formats and algorithms are well studied in computer science and electrical engineering, we refer the interested read to \cite{higham2002_fp32book}. However, these studies usually focus on high bit with floating point formats (e.g. FP32 or FP64) and do not consider the impact on neural networks which are significantly more robust to noise.

Early work by \cite{wang2018_fp8_training} showed that convolutional networks can be trained using FP16 and FP8 using the 2M5E format. The work of \cite{sun2019_HFP8} introduces a hybrid FP8 format and first discovered that 3M4E has better inference performance than 2M5E. While they use a different FP8 format for the backwards path, they keep the bias and mantissa and exponent division fixed for all layers in the forward path. \cite{Huang2021_8bit_flex_fp} introduces a fully flexible FP8 format similar to the one we consider. They perform a layer-wise optimization to find the best configuration (mantissa, exponent, sign bit, bias). Their work does not go in-depth in the analysis of why some formats are better, and it has a fairly limited results section with easy to quantize models like ResNet18 and VGG \cite{krishnamoorthi2018quantizing}.
\cite{sun2020_4bit_training} uses a custom FP4 format for the backwards path and INT4 for the forward path to enable full 4 bit training with a small drop in performance. 

To the best of our knowledge we are the first work with an extensive study of the different FP8 formats based on both analytical insights and empirical results across several tasks and data modalities. As opposed to our new FP8 simulation method, other works rely on dedicated FP8 simulations that do not allow for efficient gradient based learning of the bias or mantissa/exponent bit configuration.

%% file: limitations.tex
\section{Impact and Limitations}
\paragraph{Impact}
FP8 is becoming widespread\footnote{\url{http://bit.ly/3Sd8wey}} as an alternative to INT8, with multiple chips from vendors like Nvidia, Intel, Graphcore, AMD and IBM moving to support 3M4E and/or 2M5E FP8 formats, often with fixed bias values. 
In this paper we show that these formats are not optimal and that many networks benefit from FP8 formats with more mantissa bits and bias values that can be flexibly set per channel.
We hope that the results in this paper will help guide FP8 hardware design decisions in the broader community.

\paragraph{Limitations}

In this paper, we restrict ourselves to studying the impact of various FP8 formats on model accuracy and ignore hardware implementation-specific impact on power consumption and latency.
Assessing the difference in hardware impact between INT8 and FP8 for all networks is not trivial. From a data transfer bandwidth perspective, the two 8-bit formats incur similar overhead, while for compute limited models, the relative overheads depend on the exact implementation. Generally, FP8 units use more power for additions and multiplications~\cite{horowitz}, however, surrounding logic (e.g. accumulator design, NaN/overflow checks etc) might amortize this difference, and in multi-purpose designs which support not only FP8, but e.g., also FP16/32, INT8/16 the difference in the overheads might disappear. Therefore we limited this work to comparing the different formats on a bit width level. Hardware design teams use our accuracy analysis to make trade-offs on the hardware side for their specific use-case and design.

%% file: conclusion.tex
\section{Conclusion}

In our analysis of the FP8 format and its benefits for quantized neural network inference we have touched on many points. We have shown that analytically the FP8 format can improve on the INT8 format for Gaussian distributions that are common in neural networks, and that higher exponent bits work well when outliers occur. This paper introduced a new way to simulate FP8 quantization in FP32 hardware that speeds up FP8 quantization simulation and makes it possible to learn the bias and mantissa-exponent bit-width trade-off. We validated the FP8 format for many networks in a post-training quantization setting, showing that generally for neural networks the 5M2E and 4M3E FP8 format works the best, and that for networks with more outliers like transformers increasing the number of exponent bits works best. We have also shown that when doing quantization-aware training, many of these benefits of the format disappear, as the network learns to perform well for the INT8 quantization grid as well. 

%% file: appendix.tex
\appendix

\section{Analytical computation of the expected quantization error}
\subsection{Input quantization error.}
\label{sec:integral_analyt}

Equation~\ref{eq:rounding_error} (the quantization error)  can be split into two terms corresponding to the rounding error $E_{rw}$ and the clipping error $E_{cw}$:

\begin{equation}
{\mathbb E}(W-Q_{\alpha}(W))^2={\mathbb E}(R_{\alpha}(W))^2=E_{rw}+E_{cw},
\end{equation}

\begin{equation}
E_{rw}=\int \limits_{\alpha_{min}}^{\alpha_{max}}R_{\alpha}^2(w)p_w(w)dw,    
\end{equation}

\begin{equation}
\label{eq:clipping_error_split}
E_{cw}=\int \limits_{-\infty}^{\alpha_{min}}(w-\alpha_{min})^2p_w(w)dw + \int \limits_{\alpha_{max}}^{\infty}(\alpha_{max}-w)^2p_w(w)dw.    
\end{equation}

As the weight and activation tensor values are bound, we assume that the distribution $p_w(w)$ is clipped within the interval $(w_{min},w_{max})$. Thus the clipping error can be written as: 
\begin{equation}
\begin{split}
 E_{cw}=\mathds{1}_{w_{min}<\alpha_{min}}\int \limits_{w_{min}}^{\alpha_{min}}(w-\alpha_{min})^2p_w(w)dw + \mathds{1}_{\alpha_{max}<w_{max}}\int \limits_{\alpha_{max}}^{w_{max}}(\alpha_{max}-w)^2p_w(w)dw,
 \end{split}
\end{equation}

where $\mathds{1}_{\alpha_{max}<w_{max}}$ is the indicator function. The calculation or the rounding error $E_{rw}$ can be split into two sub-intervals for each interval $(\alpha_i,\alpha_{i+1})$ where the first sub-interval corresponds to rounding up and the second sub-interval corresponds to rounding down:

\begin{equation}
\begin{aligned}
\label{eq:rounding_error_split}
 E_{rw}= \sum_{i=1}^{|\alpha|} \int \limits_{\alpha_i}^{\alpha_{i+1}}R_{\alpha}^2(w)dw= 
 \sum_{i=1}^{|\alpha|}   \int\limits_{\alpha_i}^{(\alpha_{i}+\alpha_{i+1})/2} (w-\alpha_i)^2 p_w(w)dw + \\ 
\sum_{i=1}^{|\alpha|}  \int\limits_{(\alpha_{i}+\alpha_{i+1})/2}^{\alpha_{i+1}} (\alpha_{i+1}-w)^2 p_w(w)dw.
\end{aligned} 
\end{equation}

In order to simplify the computation, we introduce the following function:
\begin{equation}
I_w(a,b,w_{0}) \coloneqq \int \limits_{a}^{b} (w-w_0)^2 p_w(w)dw.
\end{equation}
Thus we can redefine~\ref{eq:rounding_error_split} as:
\begin{equation}
E_{rw} = \sum_{i=1}^{|\alpha|} \left[ I(\alpha_i,(\alpha_{i}+\alpha_{i+1})/2,\alpha_i) + I((\alpha_{i}+\alpha_{i+1})/2,\alpha_{i+1},\alpha_{i+1})  \right].
\end{equation}

The sum of the rounding errors for each interval between two representable grid-points. We note that the clipping error $E_{cw}$ can also be expressed using $I_w(a,b,w_{0})$: 
\begin{equation}
\begin{split}
 E_{cw}= I_w(w_{min},\alpha_{min},\alpha_{min}) + I_w(\alpha_{max},w_{max},\alpha_{max}).
 \end{split}
\end{equation}

The analytical expressions for $I(w_{min},\alpha_{min},\alpha_{min})$ for different distributions are given in Appendix \ref{sec:appendix_distr_formulas}. Thus, given the explicit definition of the quantization grid and the probability density function, we can analytically compute the rounding error for different distributions, for example the Gaussian, Uniform, or Student's t-distribution. 

\subsection{Scalar product quantization error.}

In this section, we describe a practical way for analytical computation of the expected MSE of the scalar product with quantized inputs ${\mathbb E} (\Delta Y^2)$.
Practically, the covariance matrix for $W$ and $X$ is diagonally dominant therefore we can assume their independence. Thus the expected value ${\mathbb E} (\Delta Y^2)$ can further be expressed as:

\begin{equation}
{\mathbb E} (\Delta Y^2) = \int \limits_{-\infty}^{\infty} \int \limits_{-\infty}^{\infty}  \left[ xR_{\alpha}(w)+ wR_{\beta}(x) +R_{\alpha}(w)R_{\beta}(x) \right]^2 p_{x}(x) p_{w}(w) dw dx,
\end{equation}

\label{sec:appendix_scalar_product}
\begin{equation}
\label{eq:scalar_product_error}
\begin{split}
{\mathbb E} (\Delta Y^2) = \int\limits_{-\infty}^{\infty} x^2 p_x(x) dx \int\limits_{-\infty}^{\infty} R_{\alpha}^2(w)p_w(w) dw + \int\limits_{-\infty}^{\infty} w^2 p_w(w) dw \int\limits_{-\infty}^{\infty} R_{\beta}^2(x)p_x(x) dx + \\ \int\limits_{-\infty}^{\infty} R_{\beta}^2(x)p_x(x) dx \int\limits_{-\infty}^{\infty} R_{\alpha}^2(w)p_w(w) dw +
2\int\limits_{-\infty}^{\infty} wR_{\alpha}(w)p_w(w)dw \int\limits_{-\infty}^{\infty} x R_{\beta}(x)p_x(x)dx + \\
2\int\limits_{-\infty}^{\infty} R_{\beta}^2(x)p_x(x) dx \int\limits_{-\infty}^{\infty} wR_{\alpha}(w)p_w(w) dw + 2\int \limits_{-\infty}^{\infty} R_{\alpha}^2(w)p_w(w) dw \int \limits_{-\infty}^{\infty}xR_{\beta}(x)p_x(x) dx.
\end{split}
\end{equation}

We note that the first term is nothing but the rounding error on $W$ (see equation (\ref{eq:rounding_error})) weighted by a non-central second moment on $X$ which does not depend on the quantization grid. The structure of the second term is very similar while $W$ and $X$ are interchanged. As the first two terms are the only integrals of non-negative functions, in practice they become dominant and mostly determine the MSE magnitude.  

In order to compute the scalar product error analytically, we rewrite equation~(\ref{eq:scalar_product_error}) in the following form:
\begin{equation}
\begin{split}
{\mathbb E} (\Delta Y^2) = M_x E_{rw} + M_w E_{rx} + E_{rw} E_{rx} +
2E_{sw} E_{sx} + 2 E_{rw} E_{sx} + 2 E_{rx} E_{sw},
\end{split}
\end{equation}
Where every term is introduced below. $M_w$ and $M_w$ are the second non-central moments for $W$ and $X$, respectively. These terms can be computed using functions $I_w(a,b,w_0)$ and $I_x(a,b,x_0) \coloneqq \int \limits_{a}^{b} (x-x_0)^2 p_x(x)dx$:
\begin{equation}
\begin{split}
M_w=\int \limits_{w_{min}}^{w_{max}} w^2 p_w(w) dw=I_w(w_{min},w_{max},0),\\
M_x=\int \limits_{x_{min}}^{x_{max}}x^2 p_x(x) dx=I_x(x_{min},x_{max},0);
\end{split}
\end{equation}
$E_{rw}$ and $E_{rx}$ are rounding errors on $W$ and $X$ similar to the expression in equation (\ref{eq:rounding_error}):
\begin{equation}
\begin{split}
E_{rw}=\int \limits_{w_{min}}^{w_{max}}R_{\alpha}^2(w)p_x(w) dw,\\
E_{rx}=\int \limits_{x_{min}}^{x_{max}}R_{\beta}^2(x)p_x(x) dx;
\end{split}
\end{equation}
finally, $E_{sw}$ and $E_{sx}$ are the following integrals.
\begin{equation}
\begin{split}
E_{sw}=\int \limits_{w_{min}}^{w_{max}} wR_{\alpha}(w)p_w(w)dw,\\
E_{sx}=\int \limits_{x_{min}}^{x_{max}} xR_{\beta}(x)p_x(x)dx.
\end{split}
\end{equation}

Similar to rounding error calculation in equation~\ref{eq:rounding_error_split}, the computation of $E_{sw}$ and $E_{sx}$ can be split into sub-intervals. For $E_{sw}$:

\begin{equation}
\label{eq:E_s_split}
 E_{sw}= \sum_{i=1}^{|\alpha|} \left[ \int \limits_{\alpha_i}^{(\alpha_{i}+\alpha_{i+1})/2} w(w-\alpha_i) p_w(w)dw + \int \limits^{\alpha_{i+1}}_{(\alpha_{i}+\alpha_{i+1})/2} w(\alpha_{i+1}-w) p_w(w)dw \right].
\end{equation}

We define $J_w(a,b,w_{0})$ as follows:
\begin{equation}
J(a,b,w_{0}) \coloneqq \int \limits_{a}^{b} w(w-w_0) p(w)dw.
\end{equation}

Thus we can express the term $E_{sw}$ in equation~\ref{eq:E_s_split} as:
\begin{equation}
E_{sw} = \sum_{i=1}^{|\alpha|} \left[ J(\alpha_i,(\alpha_{i}+\alpha_{i+1})/2,\alpha_i) - J((\alpha_{i}+\alpha_{i+1})/2,\alpha_{i+1},\alpha_{i+1})  \right].
\end{equation}
The term $E_{sx}$ can be computed in a similar way. The formulas for $J_w(a,b,w_{0})$ for different distributions are given in Appendix \ref{sec:appendix_distr_formulas}.

\subsection{Formulas for integrals $I_x(a,b,x_0)$ and $I_w(a,b,x_0)$.}
In this section we give formulas for the functions $I_x(a,b,x_{0})$ and $J_x(a,b,x_{0})$ for different distributions which are necessary for the analytical computation of the rounding error and the scalar product error. The formulas for $I_w(a,b,w_{0})$ and $J_w(a,b,w_{0})$ are similar while $p_w(w)$ is used as the probability density function.
The formulas were obtained using symbolical computations.

\label{sec:appendix_distr_formulas}
\textbf{Gaussian distribution.}
\begin{equation}
p(x) = \frac{1}{Z}\exp{\left[-\frac{1}{2}\left({\frac{x-\mu}{\sigma}}\right)^2\right]},    
\end{equation}

where
\begin{equation}
Z=\frac{1}{\sigma\sqrt{2\pi}},
\end{equation}

\begin{equation*}
\begin{aligned}
I_x(a,b,x_0)=- \exp \left( \frac{-a^2/2 + a\mu - \mu ^2/2}{\sigma^2} \right) \sigma^2 (-a - \mu + 2x_0)/Z + \\ \sqrt{\frac{\pi}{2}}\sigma(-\mu^2 - \sigma ^2 + 2 \mu x_0 - x_0^2)  \erf \left[\frac{-a + \mu}{\sigma \sqrt{2}} \right]/Z + \\ \exp \left[\frac{-b^2/2 + b \mu - \mu^2/2} {\sigma^2}\right] \sigma^2 (
                -b - \mu + 2 x_0)/Z + \\
                \sigma \sqrt{\frac{\pi}{2}}(-\mu^2 - \sigma^2 + 2\mu x_0 - x_0^2) \erf \left[\frac{-b + \mu}{\sigma \sqrt{2}}\right]/Z,
\end{aligned}                    
\end{equation*}

\begin{equation}
\begin{aligned}
J_x(a,b,x_0)=x_0 \sigma\exp \left[-\frac{\mu^2}{2\sigma^2}
\right]\exp\left[\frac{(-a/2+\mu)a}{\sigma^2}\right]/Z-\\
x_0 \sigma\exp \left[-\frac{\mu^2}{2\sigma^2}\right]\left(\exp\left[\frac{(-b/2+\mu)b}{\sigma^2}\right]\sigma-\sqrt{\frac{\pi}{2}} \mu
                           \erf\left[\frac{a-\mu}{\sqrt{2}\sigma}\right]+\sqrt{\frac{\pi}{2}} \mu \erf\left[\frac{b-\mu}{\sqrt{2}\sigma}\right]\right)/Z - \\
                          \frac{\sigma}{Z}\left(\exp\left[\frac{-a^2/2+a\mu-\mu^2/2}{\sigma^2}\right]\sigma(a+\mu) +\sqrt{\frac{\pi}{2}}(\mu^2-\sigma^2)\erf\left[\frac{\mu-a}{\sqrt{2}\sigma}\right]\right) + \\
                          \frac{\sigma}{Z}\left(\exp\left[\frac{-b^2/2+b\mu-\mu^2/2}{\sigma^2}\right]\sigma(b+\mu) +\sqrt{\frac{\pi}{2}}(\mu^2-\sigma^2)\erf\left[\frac{\mu-b}{\sqrt{2}\sigma}\right] \right).
\end{aligned}                           
\end{equation}
\textbf{Uniform distribution.}

\begin{equation}
p(x) =\begin{cases}
			p_0, & \text{if $a\leq x \leq b$}\\
            0, & \text{otherwise},
		 \end{cases}
\end{equation}

\begin{equation}
I_x(a,b,x_0)= p_0\left(-\frac{a^3+ b^3}{3} + (a^2 - b^2) x_0 + (b-a) x_0 ^2 \right),
\end{equation}

\begin{equation}
J_x(a,b,x_0)= \frac{a^2-b^2}{2}p_0+(b-a)p_0x_0.    
\end{equation}

\textbf{Student's t-distribution.}

\begin{equation}
p(x) = \frac{1}{Z} \left(1+\frac{t^2}{\nu}\right)^{-(\nu+1)/2},
\end{equation}

\begin{equation}
Z=\sqrt{\nu} \mathcal{B}\left(\frac{1}{2},\frac{\nu}{2}\right),    
\end{equation}

\begin{equation}
\begin{aligned}
I_x(a,b,x_0) = \frac{2\nu x_0(-1+(\frac{a^2+\nu}{\nu}))^{\frac{1-\nu}{2}}}{(1-\nu)Z} - 
\frac{2\nu x_0(-1+(\frac{b^2+\nu}{\nu}))^{\frac{1-\nu}{2}}}{(1-\nu)Z} +
\\
-a u^2 \mathrm{hyp2f1}(1/2,(1+\nu)/2,3/2,-a^2/\nu)/Z + \\ b u^2 \mathrm{hyp2f1}(1/2,(1+\nu)/2,3/2,-b^2/\nu)/Z - \\
a^3 \mathrm{hyp2f1}(3/2,(1+\nu)/2,5/2,-a^2/\nu)/(3Z) + \\ b^3 \mathrm{hyp2f1}(3/2,(1+\nu)/2,5/2,-b^2/\nu)/(3Z),
\end{aligned}
\end{equation}

where $\mathrm{hyp2f1}(a,b,c,d)$ is Gaussian hypergeometric function.

\begin{equation}
J_x(a,b,x_0)= \frac{\nu ^ {(1 + \nu) / 2}x_0}
{1 - \nu}\left[-(a ^ 2 + \nu)^{(1 - \nu)/2} + (b^2 + \nu) ^ {(1 - \nu) / 2}\right]/Z +I_x(a,b,0).
\end{equation}

\section{Ablations}
\subsection{Quantization error ablation}
\label{sec:appendix_conv_layer_ablation}

\begin{center}
\begin{figure}[t]
\centering
\begin{small}
\begin{tabular}{c}
\includegraphics[width=6.0cm]{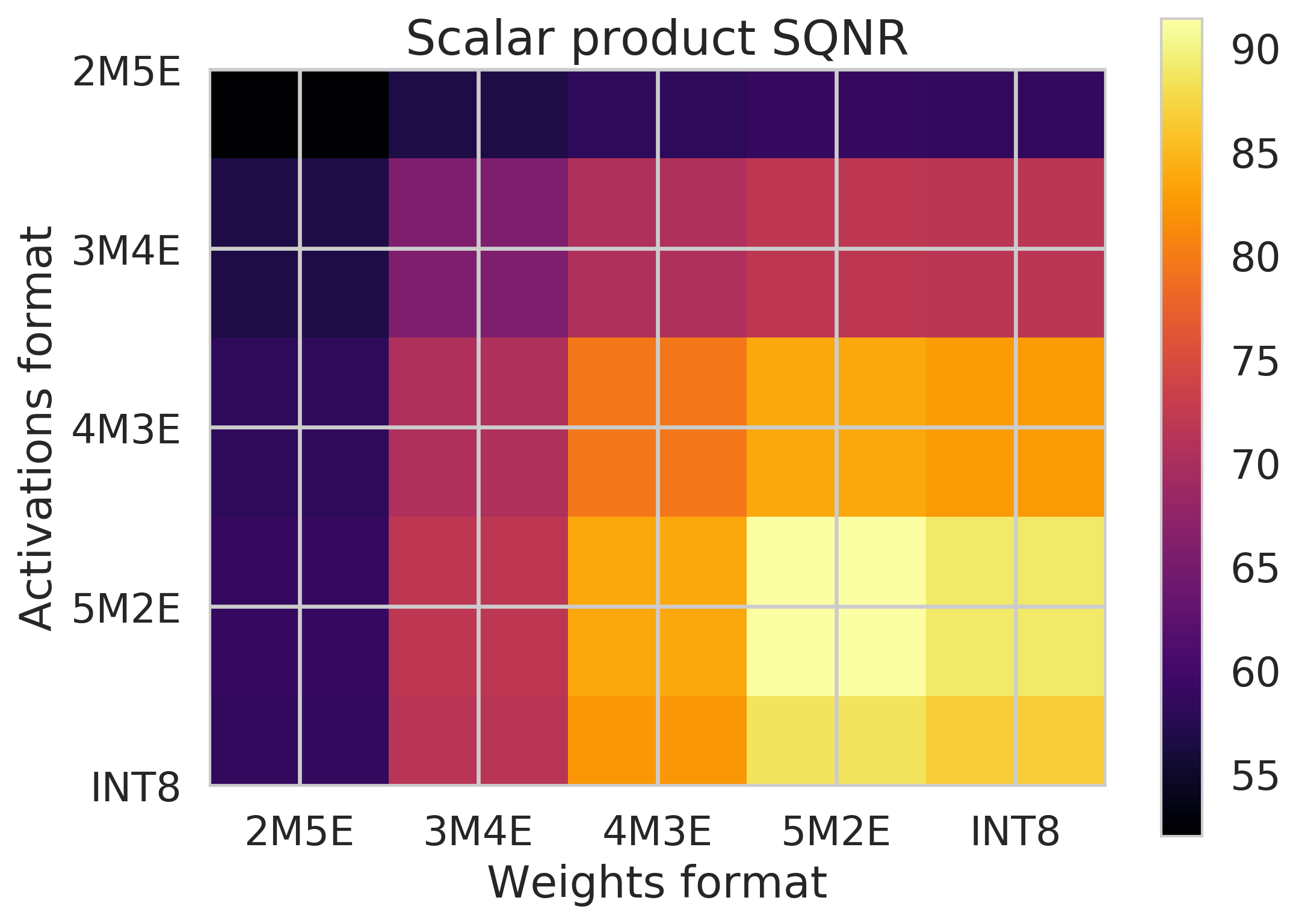} 
\end{tabular}
\end{small}
\caption{Study of a layer of pre-trained Resnet18 model. We fit two Gaussian distributions into weights and activations sample. Weights $W\sim \mathcal{N}(-1.0\times10^{-3},1.7\times10^{-2})$, clipped at $[-0.35,0.35]$, activations $X\sim \mathcal{N}(0.06, 0.11)$ clipped at $[0,3.63]$. The optimal format is M5E2 for both weights and activations.
}
\label{fig:resnet18_conv_2d_sqnr}
\end{figure}
\end{center}

In this section we analyze different combinations of floating point formats for weights and activations based on the analytical computation of the scalar product error. We consider two Gaussian distributions fit into weights and activations distributions of a layer of pre-trained Resnet18 model. The results are shown in fig.~\ref{fig:resnet18_conv_2d_sqnr}. We observe that the optimal format for both weights and activations is 5M2E. In order to facilitate visual comparison of MSE values of different magnitude we plot SQNR which is defined as follows:
\begin{equation}
10\log_{10}\left(\frac{\mathbb{E}\left[\mW^2\right]\mathbb{E}\left[\mX^2\right]}{\mathbb{E}\left[\left(\mW\mX - Q_{\alpha}(\mW)Q_{\beta}(\mX)\right)^2\right]}\right).
\end{equation}

\subsection{Comparison of the analytical and empirical rounding error.}

\begin{center}
\begin{figure}[t]
\centering
\begin{small}
\begin{tabular}{cc}
\includegraphics[width=6.5cm]{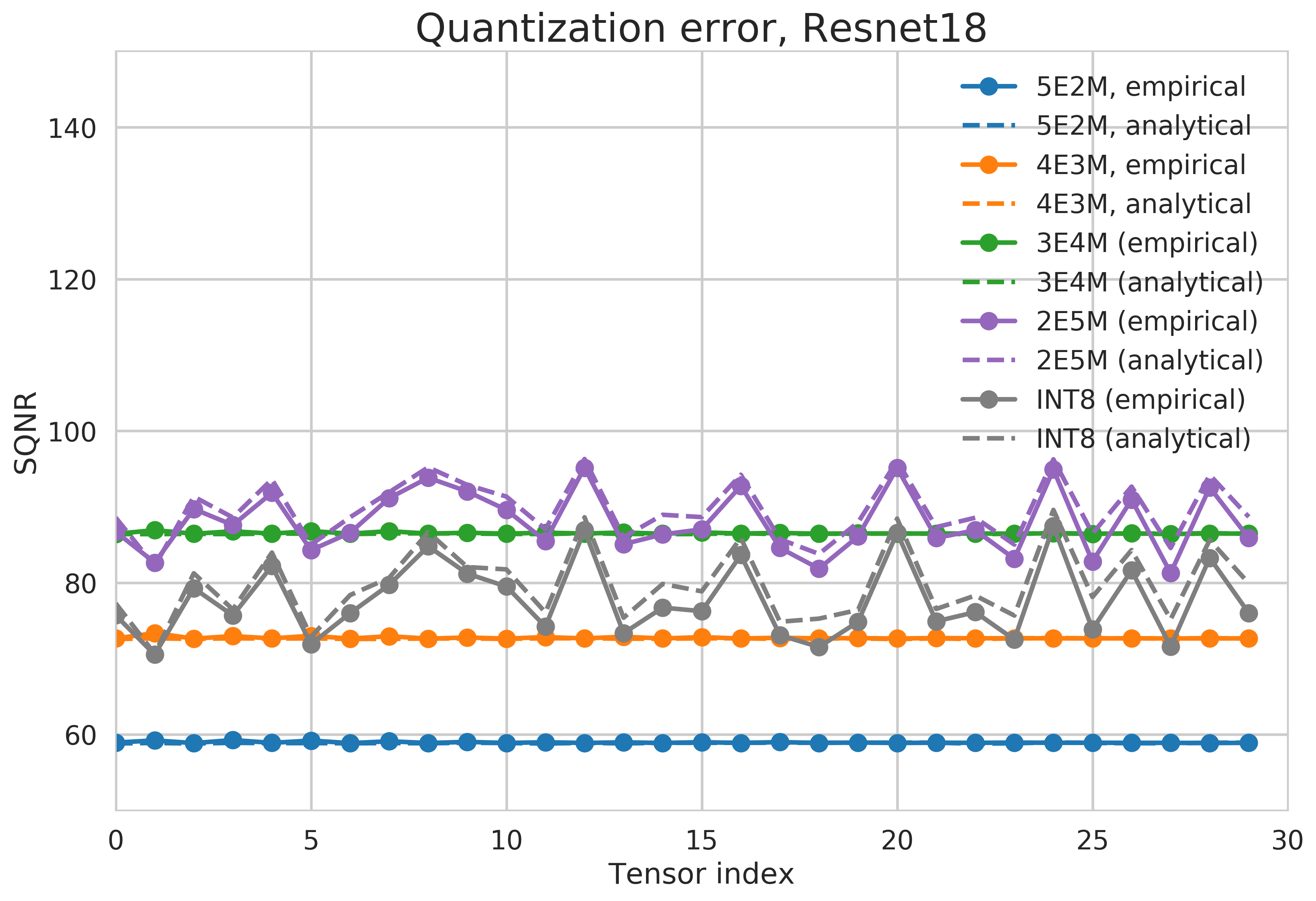} & \includegraphics[width=6.5cm]{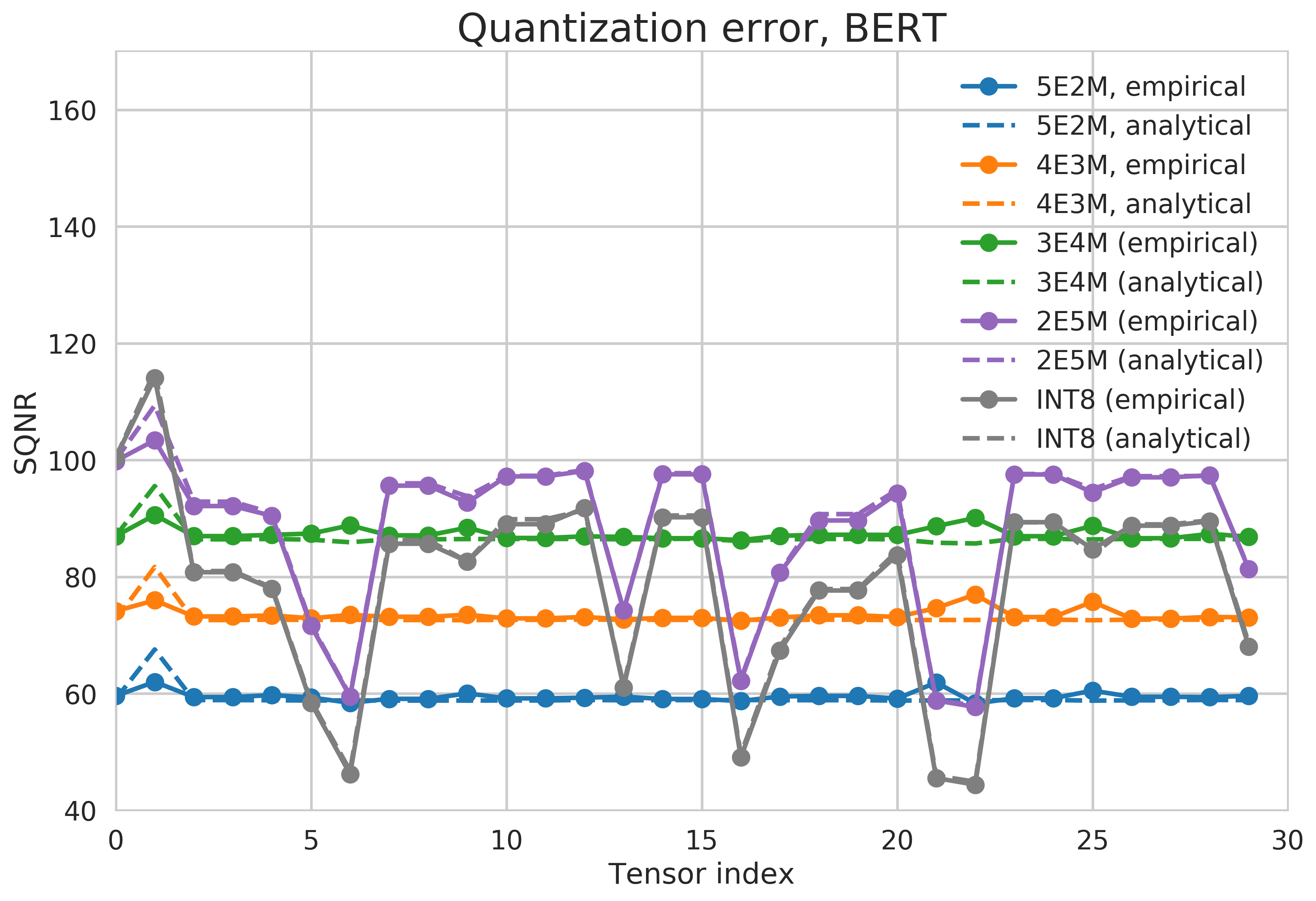}  
\end{tabular}
\end{small}
\caption{Comparison of the expected quantization error computed analytically to empirical quantization error. A random subset of 30 weights/activations tensor is chosen for Resnet18 and Bert.}.

\label{fig:per_tensor_sqnr}
\end{figure}
\end{center}

In this section we compare the expected rounding error computed analytically to the empirical rounding error. The results are given on fig.~\ref{fig:per_tensor_sqnr} Note that, for visual purposes, we plot the signal-to-quantization-noise ratio (SQNR) instead of MSE in these plots.
SQNR is log-proportional to MSE; i.e. a value that minimizes MSE will maximize SQNR. 
SQNR for an input tensor $\mX$ is defined as:
\begin{equation}
10\log_{10}\left(\frac{\mathbb{E}\left[\mX^2\right]}{\mathbb{E}\left[\left(\mX - Q_{\alpha}(\mX)\right)^2\right]}\right).
\end{equation}

\section{Examples of tensors in Bert}
\begin{figure}[h]
\centering
\begin{tabular}{cc}
\includegraphics[height=4.0cm]{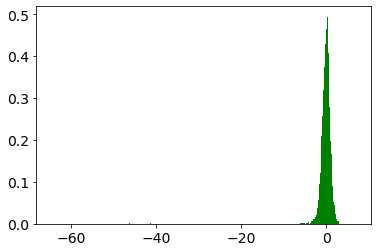} & \includegraphics[height=4.0cm]{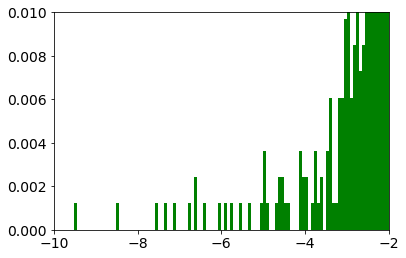} \\
(a) & (b) \\
\includegraphics[height=4.0cm]{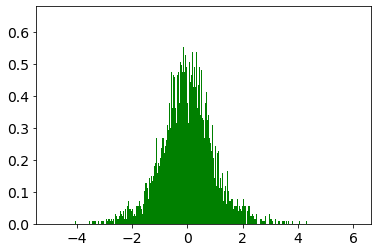} & \includegraphics[height=4.0cm]{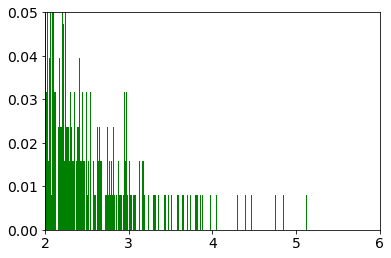} \\
(c) & (d) 
\end{tabular}
\caption{Examples of tensors in the forward pass of Bert. The plots (b) and (d) are scaled versions of the plots (a) and (b) respectively. The tails in the distributions contain significant amount of outliers.}
\label{fig:bert_tensors_examples}
\end{figure}

In this section we give examples of strong outliers in the tensors in Bert (fig.~\ref{fig:bert_tensors_examples}).

\section{Importance of the outliers}

\begin{center}
\begin{figure}[h]
\centering
\begin{small}
\begin{tabular}{c}
\includegraphics[width=8.0cm]{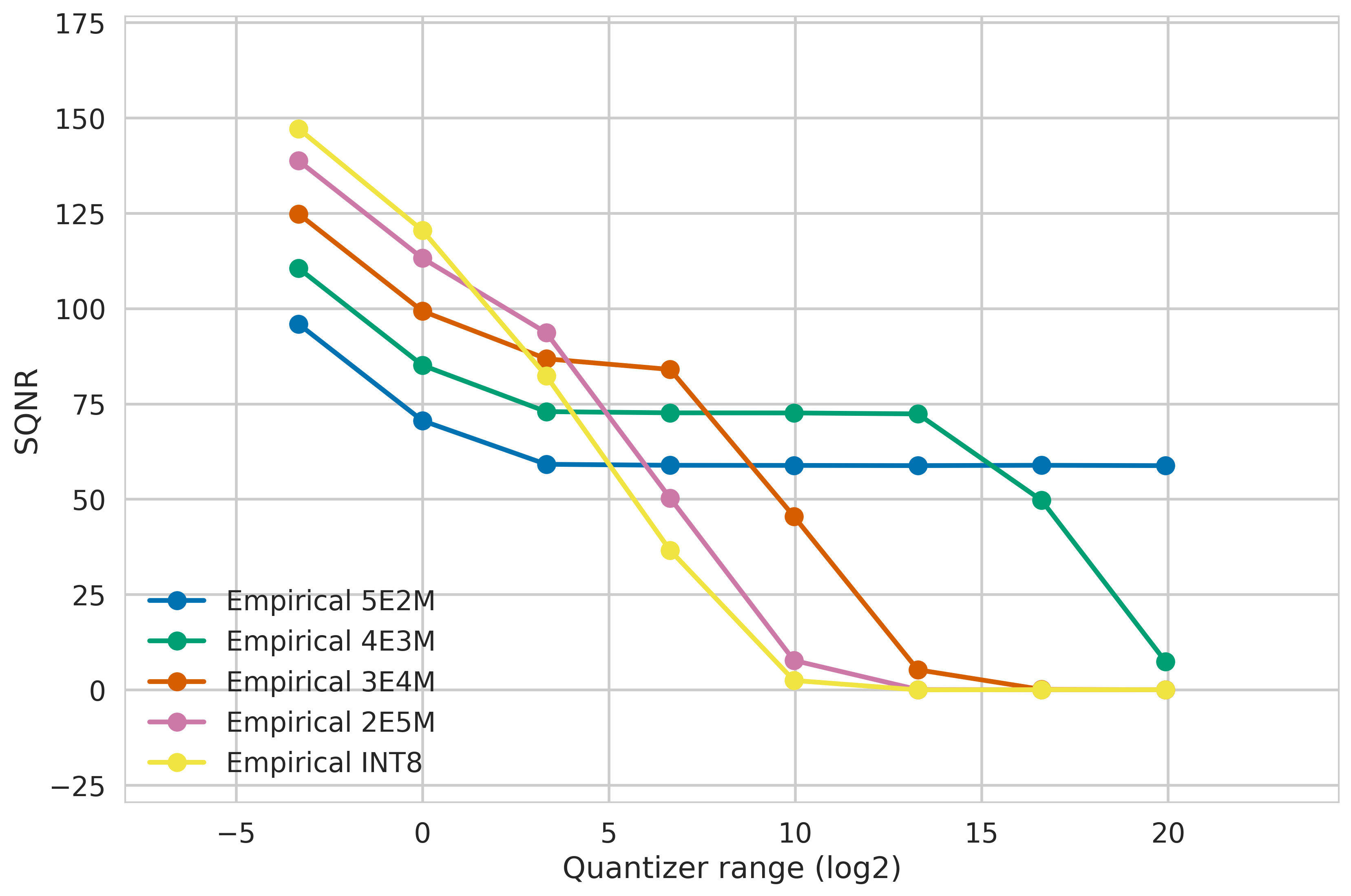} 
\end{tabular}
\end{small}
\caption{SQNR for Student's-t distribution with $\nu=2$ for different quantizer range. Optimal exponent bit-width values increases while the quantization range is being expanded.
}
\label{fig:students_t_outliers}
\end{figure}
\end{center}

In this ablation we demonstrate influence of the outliers on the choice of the optimal floating point format. The experiment is based on the analytical computation of the rounding error. We consider a Student's-t distribution with $\nu=2$ with increasing quantization range. Min-max quantizer range estimator is used, the distribution is clipped at the quantization range. The results are given on fig.~\ref{fig:students_t_outliers}. While the quantization range is being increased, the optimal exponent bit-width values grows starting from zero (INT8 format) to 5-bit exponent.

\section{MSE-based mantissa bits and bias}\label{sec:mse_mantissa_bias}

In our quantization setup, each tensor has an individual quantizer.
The quantizers can use per-tensor or per-channel quantization.
In case per-channel quantization is used, each channel has its own clipping value $c$ (and thus its own value for $\bhat$), while the number of mantissa bits is set for a full tensor, and is thus shared across all channels.

During quantizer initialization we use the input weight tensor or one batch of activations, both referred to as $\mX$, and find values for $m$ and $c$ that minimize reconstruction MSE.
In order to do so, we perform a grid search over values of $m$ and values of $c$.
For $m$ we consider 1, 2, 3, 4, 5, and 6 bits. 
For $c$ we find the absolute maximum value $\sigma$ of $\mX$ (or, in case of per-channel quantization, each channel in $\mX$), and run the grid search over 111 evenly spaced values between $0.1\sigma$ and $1.2\sigma$.
We then record the values of $m$ and $c$ that minimize MSE and initialize the quantizer with these values.

\begin{figure}[h]
    \centering
    \includegraphics[width=\textwidth]{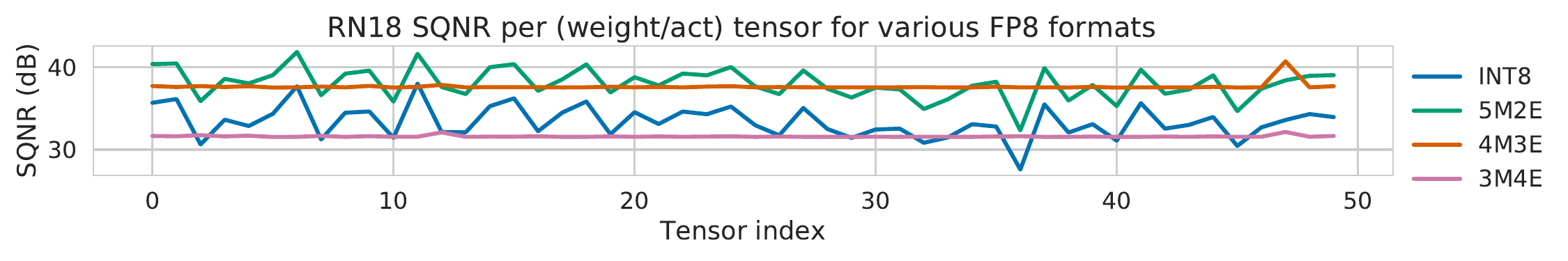}
    \caption{SQNR per layer for ResNet18.}\label{fig:rn18sqnr}
\end{figure}

\begin{figure}[h]
    \centering
    \includegraphics[width=\textwidth]{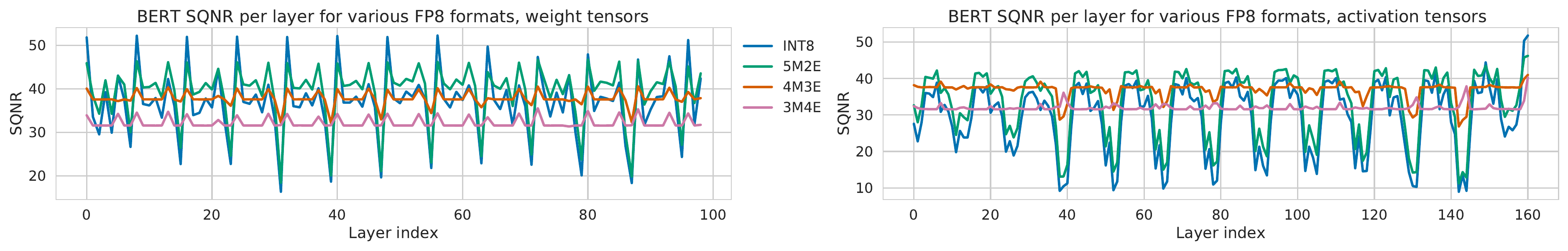}
    \caption{SQNR per layer for BERT (split up into weights and activations). }\label{fig:bertsqnr}
\end{figure}

In case per-channel quantzation is used, we run this procedure for each channel individually.
On each channel, for each value of $m$, we store the value of $c$ that minimized quantization on that channel.
We then choose a per-tensor value of $m$ by majority vote, i.e. the value of $m$ that occurred most often over all channels.
In case of a tie we choose the value of $m$ that has lowest cumulative MSE.
Lastly, we set $c$ to the value that minimized MSE for the per-tensor value of $m$.
We also experimented with choosing $m$ based on lowest cumulative MSE directly, but found negligible difference in resulting accuracy.
We decided to use majority vote to ensure channels with relatively large magnitude to dominate the choice of $m$. Figures \ref{fig:rn18sqnr} and \ref{fig:bertsqnr} show a per-layer analysis of the bitwidth choices that minimize MSE.

\section{Correlation between MSE in the output activations and the model accuracy\label{sec:mse_accuracy_correlation}}
\begin{figure}[h]
    \centering
    \includegraphics[width=180pt]{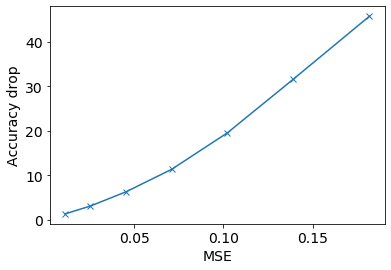}
    \caption{An ablation performed on one layer of Resnet18 pre-trained on ImageNet. Gaussian noise of increasing amplitude is added to the weights of the layer. The MSE in the output activations has strong correlation in the drop in top-1 accuracy of the model. }\label{fig:mse_accuracy_correlation}
\end{figure}
In this section we provide an example of a correlation between MSE in output activations of a layer and the full model accuracy. We take a pre-trained Resnet18 model, and consider one if its layers. We inject Gaussian noise of increasing amplitude in its weight values, and measure MSE in the output activations of the layer, and the final top-1 classification accuracy. The results are shown in figure~\ref{fig:mse_accuracy_correlation}. The MSE value and the final accuracy exhibit strong correlation, i.e. the normalized correlation coefficient value for this experiment is 0.98. 

\section{Full experimental details}\label{sec:expdetail}
For our QAT experiments, we initialized our models using the settings that gave the best results for the fixed format, the flexible bias format, and the fully flexible format.
See Table~\ref{tab:ptq} and the tables in \ref{sec:appendixptq} for details.
We then trained ResNet18 for 20 epochs, and MobileNetV2 for 10.
We used the Adam \cite{kingma2014adam} to optimize the weights. 
We considered starting learning rates of $10^{-5}$ and $10^{-6}$ as these gave best results in a pilot experiments.
Learning rates were decayed to a factor of $10^{-2}$ of the starting learning using cosine decay.

In experiments where FP8 parameters ($c$ and $m$) were learned as well, we used SGD without momentum.
We considered learning rates $10^{-2}$, $10^{-3}$, $10^{-4}$, and $10^{-5}$.
No weight decay was applied to any of the models.

Baseline INT8 QAT results followed the procedure as described in \cite{whitepaper}.

\section{Gradients of the FP8 quantizer}
As stated previously, the FP8 quantizer gives the `straight-through' gradient \cite{bengio2013estimating} w.r.t. the values to be quantized:

\begin{equation}
    \frac{\partial}{\partial x_i} F(x_i, m, c) =
    \begin{dcases}
        1, & \text{if } -c \leq x_i \leq c \\
        0, & \text{otherwise}.
    \end{dcases}
    \label{eq:grad_of_xi}
\end{equation}

The gradient w.r.t. $c$ is as follows:
\begin{equation}
    \frac{\partial}{\partial c} F(x_i, m, c) = 
    \begin{dcases}
        \frac{2^{p_i}}{c}\left(\left\lfloor \frac{x_c}{s} \right\rceil - \frac{x_c}{s} \right), & \text{if } \left\lfloor \log_2 |x_i| +\bhat \right\rfloor > 1 \text{ and } -c \leq x_i \leq c, \\
        \frac{1}{c\ln(2)}, & \text{if } \left\lfloor \log_2 |x_i| +\bhat \right\rfloor \leq 1 \text{ and } -c \leq x_i \leq c, \\
        -1, & \text{if } x_i < -c, \\
        1, & \text{if } x_i > c, \\
    \end{dcases}
    \label{eq:pi_with_scale}
\end{equation}

Lastly, the gradient w.r.t. $m$ is as follows:

\begin{equation}
    \frac{\partial}{\partial m} F(x_i, m, c) = 2^{p_i}\ln(2)\left(\left\lfloor \frac{x_c}{s} \right\rceil - \frac{x_c}{s} \right)\left(2^{7-\lfloor m \rceil}+\frac{2^{-m}}{\left(2 - 2^{-m} \right)} \right)
\end{equation}

\section{PTQ results}\label{sec:appendixptq}

The best results for INT8, FP8 with fixed bias (per Mantissa/Exponent division), FP8 with flexible bias, and fully flexible FP8, on all models considered in Section~\ref{sec:ptq} are shown in Table~\ref{tab:ptqfull}.

Per-task BERT results are included in Table~\ref{tab:bertfull}.

\input{tables/fp8_ptq_joint_table}

\input{tables/fp8_bert_tasks}

\section{DeepLabV3 weight distribution}
\begin{figure}[h]
    \centering
    \includegraphics[width=\linewidth]{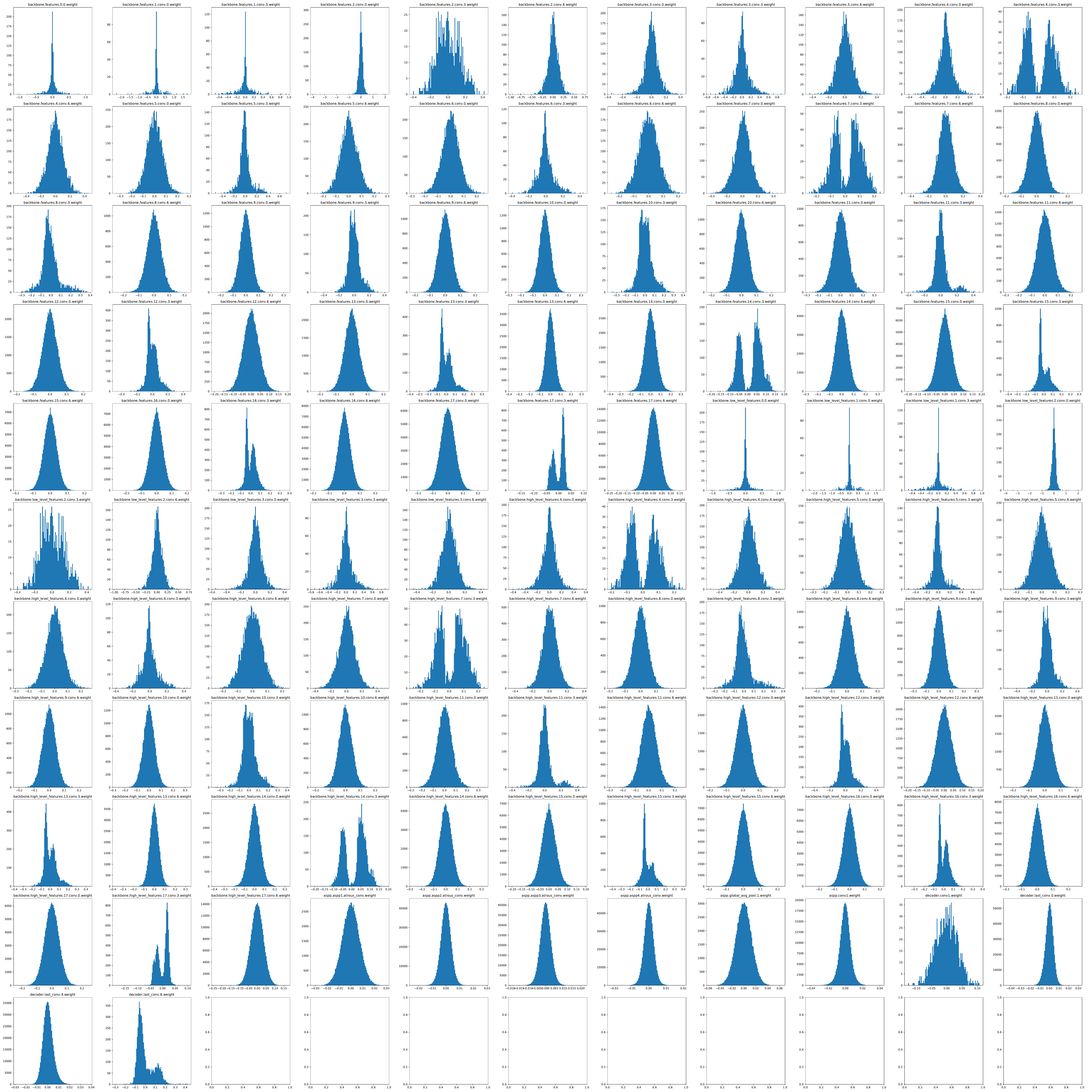}
    \caption{A plot showing the distributions for each weight tensor in the DeepLabV3 model instance used in our experiments.}\label{fig:deeplab_weights}
\end{figure}

DeepLabV3 shows a larger degradation in the 'fixed format' PTQ setting than other models considered. 
Figure \ref{fig:deeplab_weights} shows the distribution of the values in each weight tensor in DeepLabV3.
We believe the low performance on fixed format PTQ to be due to the fact that some layers in early the DeepLabV3 backbone have relatively large outliers (e.g. backbone.features.2.conv.0.weight, row 1 column 4), necessitating a relatively large number of exponent bits, while some layers later in the backbone and in the decoder have distributions that require few exponent bits (e.g. backbone.high\_level\_features.15.conv.3.weight; backbone.high\_level\_features.16.conv.3.weight; decoder.last\_conv.8.weight). 
This discrepancy is present in other networks, most notably MobileNetV2, however, it is not as prominent as in this network.

%% file: tables/fp8_ptq_joint_table.tex
\begin{table}
    \centering
        \small
        \begin{tabular}{l|rr|rrr|rrrr|r}
        
        {}              & FP32       & INT8           & \multicolumn{3}{l|}{Fixed bias} & \multicolumn{4}{l|}{Flex bias} &        Flex \\
        {}              &            &                &            5M2E &            4M3E &            3M4E &            5M2E &            4M3E & 3M4E & 2M5E & \\
        \hline
        ResNet18        &      69.72 &           69.64 &           27.25 &           68.35 &  \textbf{68.44} &  \textbf{69.66} &           69.45 &           68.57 & 64.92 &       69.64 \\
        MobileNetV2     &      71.70 &           70.94 &           20.35 &  \textbf{68.49} &           66.50 &  \textbf{71.06} &           70.62 &           66.05 & 49.51 & \textbf{71.28} \\
        ViT             &      77.75 &           76.41 &           51.05 &           76.97 &  \textbf{77.37} &           77.30 &  \textbf{77.71} &           77.56 & 76.69 &       77.62 \\
        DeepLabV3       &      72.91 &           71.24 &            3.27 &            3.27 &   \textbf{9.42} &  \textbf{72.58} &           71.28 &           37.93 & 6.53   & \textbf{72.72} \\
        BERT            &      83.06 &           71.03 &             N/A &           78.65 &  \textbf{83.06} &           80.31 &           82.61 &  \textbf{82.80} & \textbf{82.81}   &       83.02 \\
        SalsaNext       &      55.80 &           54.22 &           22.24 &           55.79 &  \textbf{54.92} &           55.52 &  \textbf{55.67} &           55.12 & 53.08   &       55.02 \\
        HRNet           &      81.05 &           80.93 &             N/A &  \textbf{80.71} &            0.48 &  \textbf{81.03} &           \textbf{81.04} &           80.77 & 80.14   & \textbf{81.06}
        
        \end{tabular}
        \caption{PTQ Results on all models. Fixed bias: best results over all fixed bias values for each Mantissa / Exponent division. Flex bias: best results over all initialization methods for each Mantissa / Exponent division. Bold numbers mark best results per model for all Fixed bias and Flex bias results, respectively. Bold numbers in the `Flexible' column indicate results where `Flexible' PTQ outperforms other PTQ methods.}
    \label{tab:ptqfull}
\end{table}

%% file: tables/fp8_bert_tasks.tex
\begin{table}
    \centering
        \begin{tabular}{l|llllllll|l}
        Format    & CoLA           & SST-2          & MRPC           & STS-B          & QQP            & MNLI           & QNLI           & RTE            & Macro avg.     \\ \hline
        FP32      & 57.27          & 93.12          & 88.36          & 89.09          & 89.72          & 84.91          & 91.58          & 70.40          & 83.06          \\ 
        INT8      & 54.74          & 92.55 & \textbf{88.53} & 81.02          & 83.81          & 50.31          & 52.32          & 64.98          & 71.03          \\ 
        5M2E flex & 58.92          & 91.86          & 87.49          & 88.50          & 89.21          & 81.69          & 80.52          & 64.62          & 80.31          \\
        4M3E flex & 56.51          & 92.32          & 88.16          & \textbf{88.99} & \textbf{89.79} & \textbf{84.87} & \textbf{91.69} & 68.59          & 82.61          \\
        3M4E flex & 57.29          & \textbf{93.12}          & 87.35          & 88.87          & 89.76          & \textbf{84.87} & 91.51          & \textbf{69.68} & \textbf{82.80} \\
        2M5E flex & \textbf{60.01} & 92.43          & 88.38          & 88.00          & 89.61          & 84.14          & 91.31          & 68.59          & \textbf{82.81}
        \end{tabular}
        \caption{Per task results for BERT for all GLUE tasks. Best quantized result for each task is marked in boldface.}
    \label{tab:bertfull}
\end{table}